\documentclass{article}

\usepackage{iclr2017_conference,times}
\usepackage{hyperref}
\usepackage{url}
\usepackage{natbib}

\usepackage[utf8]{inputenc} 
\usepackage[T1]{fontenc}    
\usepackage{url}            
\usepackage{booktabs}       
\usepackage{amsfonts}       
\usepackage{nicefrac}       
\usepackage{microtype}      
\usepackage{amsmath}
\usepackage{epsf}
\usepackage{epsfig}
\usepackage{graphicx}
\usepackage{subfigure}
\usepackage{color}
\usepackage{multirow}
\usepackage{csquotes}
\usepackage{appendix}
\usepackage[inline,shortlabels]{enumitem}

\newcommand{\argmax}{\arg\!\max}

\newcommand{\Vx}{\mathbf{x}}
\newcommand{\Vh}{\mathbf{h}}
\newcommand{\Vc}{\mathbf{c}}

\newcommand{\Ve}{\mathbf{e}}

\newcommand{\Vs}{\mathbf{s}}
\newcommand{\Vd}{\mathbf{d}}
\newcommand{\Vr}{\mathbf{r}}
\newcommand{\Vz}{\mathbf{z}}

\newcommand{\Vtheta}{\boldsymbol{\theta}}
\newcommand{\SetR}{\mathbb{R}}

\newcommand{\MW}{\mathbf{W}}
\newcommand{\MU}{\mathbf{U}}
\newcommand{\MS}{\mathbf{S}}
\newcommand{\MR}{\mathbf{R}}

\title{An Efficient  Character-Level \\ Neural Machine Translation}

\author{Shenjian Zhao \&  Zhihua Zhang   \\
  Department of Computer Science and Engineering\\
  Shanghai Jiao Tong University\\
  \texttt{\{sword.york,zhzhang\}@gmail.com} \\
}

\begin{document}

\maketitle

\begin{abstract}
Neural machine translation aims at building a single large neural network that 
can be  trained to maximize translation performance. The encoder-decoder 
architecture with an attention mechanism achieves a translation performance 
comparable to the existing state-of-the-art phrase-based systems on the task of 
English-to-French translation.  However, the use of large vocabulary becomes the bottleneck 
in both training and improving the performance. In this paper, we propose an efficient architecture
to train a deep character-level neural machine translation by introducing  a \emph{decimator} 
and an \emph{interpolator}.  The decimator is used to sample the source 
sequence before encoding while the interpolator is used to resample 
after decoding.  Such a deep model has  two major advantages.  It avoids
the large vocabulary issue radically; at the same time, it is much faster and 
more memory-efficient in training than conventional character-based models.
More interestingly, our model is able to 
translate the misspelled word like human beings. 
\end{abstract}

\section{Introduction}

Neural machine translation (NMT) attempts to build a single large neural network 
that reads a sentence and outputs a translation \citep{kalchbrenner2013recurrent, 
cho2014properties, sutskever2014sequence}. Most of the extant neural 
machine translations models belong to a family of word-level encoder-decoders 
\citep{sutskever2014sequence, cho2014learning}. 
\citet{bahdanau2014neural}  recently proposed a model with 
attention mechanism which automatically searches the alignments and greatly improves the performance.
However, the use of a large vocabulary seems necessary 
for the word-level neural machine translation models to improve performance 
\citep{sutskever2014sequence, chousing}.

\citet{chung2016character}  listed three reasons behind the wide adoption 
of word-level modeling:  
\begin{enumerate*}[(i)]
  \item word is a basic unit of a language,
  \item data sparsity,
  \item  vanishing gradient of character-level modeling.
\end{enumerate*}
Considering that a language itself is an evolving system,  
it is impossible to cover all words in the language.  The 
problem of rare words that are out of vocabulary (OOV) is an important issue which 
can effect the performance of neural machine translation. Using larger 
vocabulary does improve performance \citep{sutskever2014sequence, chousing}, but 
the training becomes much harder and the vocabulary is often filled with many 
similar words that share a lexeme but have different morphology.  

There are many approaches to dealing with the out-of-vocabulary issue.  
\citet{gulcehre2016pointing, luong2014addressing, chousing} proposed to obtain
the alignment information of target unknown words, after 
which simple word dictionary lookup or identity copy can be performed to replace 
the unknown words in  translation.   These approaches ignore several important 
properties of languages such as monolinguality and crosslinguality as pointed 
out by  \citet{luong2016achieving}.  \citet{luong2016achieving} further 
proposed a hybrid neural machine translation model which leverages the power of 
both words and characters to achieve the goal of open vocabulary neural machine 
translation.  

Intuitively, it is elegant to directly model pure characters. However, as the length of 
sequence grows significantly,  character-level translation models have 
failed to produce competitive results compared with word-based models. In 
addition, they require more memory and computation resource. Especially,  it 
is much difficult  to train the attention  component. \citet{ling2015finding} 
proposed a compositional character to word (C2W) model and applied to machine 
translation in \citep{ling2015character}.  However, they found that it is slow 
and difficult to train source character-level models and had to resort to layer-wise
training the neural network and applying supervision for the 
attention component.

In order to address the training issue mentioned in  \citep{ling2015character},
we introduce a \emph{decimator} before encoding as well as an  \emph{interpolator} after decoding. 
The decimator is  based on a variant of the gate recurrent unit 
(GRU) \citep{cho2014learning, chung2014empirical},  which samples the character 
sequence according to the occurrence of delimiter (usually the space) and resets 
to the initial state accordingly. The interpolator also relies on a 
variant of GRU, which sets the state to the output of decoder and generates 
character sequence until generating a delimiter. In this way,  we almost keep 
the same encoding length for encoder as word-based models but eliminate the use 
of a large vocabulary. Besides, the decoding step is much more natural compared 
with \citep{bahdanau2014neural} which uses a multi-layer network following a 
softmax function to compute the probability of each target word. Furthermore, we are able
to efficiently train the deeper model which using multi-layer recurrent network and achieve higher
performance.


In summary, we propose an efficient architecture to train a deep character-level 
neural machine translator. We  show that the model achieves a higher translation 
performance compare to  the word-level neural machine translation on the task of English-to-French
translation. Furthermore, our model is able to translate the misspelled words 
and learn similar embeddings of the words with similar meanings.


\section{Neural Machine Translation}
\label{sec:background}

Neural machine translation is often implemented as an encoder-decoder 
architecture. The encoder usually uses a recurrent neural network (RNN) or a 
bidirectional  recurrent neural network (BiRNN)  
\citep{schuster1997bidirectional} to encode the  input sentence $\Vx = \{x_1, 
\ldots,x_T\}$ into a sequence of hidden states $\Vh=\{\Vh_1, \ldots, \Vh_T\}$:
\begin{align*} 
 \Vh_t = f_1(\Ve(x_t), \Vh_{t-1}),
\end{align*}
where $\Ve(x_t) \in \SetR^m$ is an m-dimensional embedding of $x_t$.
The decoder, another  RNN, is often trained to predict next word 
$y_t$ given previous predicted words $\{ y_1, \ldots, y_{t-1} \}$ and the 
context vector $\Vc_t$:
\begin{align*}
p(y_t | \{y_1, \ldots, y_{t-1} \}) = g(\Ve(y_{t-1}), \Vs_t, \Vc_t),
\end{align*}
where 
\begin{align*}
\Vs_t = f_2(\Ve(y_{t-1}), \Vs_{t-1}, \Vc_t ) 
\end{align*}
and $g$ is a nonlinear and potentially multi-layered function that computes the 
probability of $y_t$. The context $\Vc_t$  depends on the sequence of 
$\{\Vh_1, \ldots, \Vh_T\}$. 
\citet{sutskever2014sequence} encoded all information in the source sentence 
into a fixed-length vector, i.e.,  $\Vc_t = \Vh_T$. \citet{bahdanau2014neural} 
computed $\Vc_t$ by the alignment model which solves the bottleneck that the 
former approach meets.

The whole model is jointly trained to maximize the conditional log-probability 
of the correct translation given a source sentence with respect 
to the parameters of the model:
\begin{align*}
\Vtheta^* = \argmax_{\Vtheta} \sum_{t=1}^T \text{log} ~p(y_t | \{ y_1,  \ldots, 
y_{t-1} 
\}, \Vx).
\end{align*}
For the detailed description of the implementation,
we refer the reader to the papers \citep{cho2014properties, 
sutskever2014sequence, bahdanau2014neural}.


\section{Deep Character-Level Neural Machine Translation}
\label{headings}

There are two problems in the word-level neural machine translation models. 
First,   how can we map a word to a vector? It is usually done by a lookup 
table (embedding matrix) where the size of vocabulary is limited. Second, how 
do we  map a vector to a word when predicting?  It is usually done via a softmax 
function. However, the large vocabulary will make the softmax intractable 
computationally.

\citet{ling2015finding,ling2015character} proposed the C2W and V2C components to address 
these two problems, however, these components are less efficient. We correspondingly devise two 
novel operators that we call  \emph{decimator} and \emph{interpolator}.
Accordingly, we propose a deep character-level neural machine translation 
model.

The decimator samples the input character sequence based on a delimiter (usually 
the space), which significantly reduces the length of input sequence. 
Thus the input of our bidirectional RNN encoder has the same length as the 
word-level encoder.  The interpolator then takes the output of decoder to 
generate a sequence of characters ending with a delimiter. This further reduces 
the burden of generating process.

\subsection{Decimator}

We introduce a variant of the gate recurrent unit (GRU) \citep{cho2014learning, 
chung2014empirical} that used in decimator  and we denote it as DGRU  (It is 
possible to use the LSTM \citep{hochreiter1997long} units  instead of the GRU described here).   DGRU reads the sequence character by 
character. Once  DGRU meets a delimiter, it will 
reset the state to the trainable initial state. 
Formally, given the input character sequence $\{ x_1, \ldots, x_t \}$, we 
first construct an auxiliary  sequence $\{ a_1, \ldots, a_t \}$ which only 
contains 0 and 1 to indicate whether $x_i$ is a delimiter.
DGRU computes the state sequence $\{ \Vh_1, \ldots, \Vh_{t+1} \}$ by 
iterating the following updates:
\begin{align}
 \Vr^j_t &= \sigma([\MW_r \Ve(x_t)]^j + [\MU_r \Vh_{t-1}]^j)  
\label{eq:reset_dgru}, \\
 \Vz^j_t &= \sigma([\MW_z \Ve(x_t)]^j + [\MU_z \Vh_{t-1}]^j), \label{eq:z_dgru} 
\\
 \tilde{\Vh}^j_t &= \phi([\MW \Ve(x_t)]^j + [\MU(\Vr_t \odot \Vh_{t-1})]^j), 
\label{eq:th_drgu} 
\\
 \hat{\Vh}^j_t &= \Vz^j_t \Vh^j_{t-1} + (1-\Vz^j_t)\tilde{\Vh}_t^j, 
\label{eq:h_dgru} \\
 \Vh_t &= (1 - a_t)  \hat{\Vh}_t   + a_t \Vh_0,
\end{align}
where $\Vh^j_t$ is the $j$-th hidden unit  of time $t$, $\Vh_0$ is the trainable
initial state, $\sigma$ is the sigmoid function and $\phi$ is the activation 
function. Note that Steps \eqref{eq:reset_dgru} to \eqref{eq:h_dgru} are the 
same as the conventional GRU \citep{cho2014learning, chung2014empirical}. The 
only difference is that  $\Vh_t$ will set to the trainable initial state $\Vh_0$ 
once a delimiter is met.

In our model, we regard the state of DGRU before  it reads a delimiter 
as  the summary of the previous character sequence (the previous word). Thus  
we need to sample $\{ \Vh_1, \cdots, \Vh_t \}$ which is the output of 
DGRU. In order to make the training more efficient, we construct a sampling 
matrix according to the delimiter in the source sequence. The sampling matrix $\MS$ has 
$t$ rows and $c$ columns, and  $c$ is the number of the delimiters. $\MS[i{-}1, j]$ is set 
to $1$ if the $j$-th delimiter is the $i$-th character of the source sequence. 
 For example, if the input character sequence is 
``\underline{g} \underline{o} \underline{ } \underline{!} \underline{ }'' and 
the output of DGRU is $[\Vh_1, \Vh_2, \Vh_3, \Vh_4, \Vh_5]$,  then the 
corresponding sample step will be:
\begin{align}
 [\Vh_1, \Vh_2, \Vh_3, \Vh_4, \Vh_5]
 \begin{bmatrix}
    0      & 0  \\
    1      & 0 \\
    0      & 0 \\
    0      & 1 \\
    0      & 0 \\
\end{bmatrix}
=
[\Vh_2, \Vh_4]. \nonumber 
\end{align}
After sampling, $[\Vh_2, \Vh_4]$ becomes the output of the decimator, thus the 
length of the sequence that needs to be encoded is significantly reduced,  which 
can be handled efficiently by the bidirectional RNN encoder.

\subsection{Interpolator}

Our deep  character-level neural machine translation also contains a variant of 
GRU in the interpolator that we call it  IGRU. IGRU has a settable state and 
generates character sequence based on the given state until generating a 
delimiter.  In our model, the state is initialized by the output of the decoder. 
Once IGRU generates a delimiter, it will set the state to the next output of the 
decoder. Given the previous output character sequence $\{y_0, y_1, 
\ldots, y_{t-1} \}$ where $y_0$ is a token representing the 
start of sentence, and the auxiliary sequence $\{a_0, a_1, \ldots,  a_{t-1} \}$ 
which is the same as decimator ($a_0$ is set to $1$), IGRU updates the state as 
follows:
\begin{align}
\Vh_{t-1} &= (1 - a_{t-1}) \Vh_{t-1} + a_{t-1} \Vd_{i_t} \label{eq:seed_state}, 
\\
\Vr^j_t &= \sigma([\MW_r \Ve(y_{t-1})]^j + [\MU_r \Vh_{t-1}]^j),  
\label{eq:reset_igru} \\
 \Vz^j_t &= \sigma([\MW_z \Ve(y_{t-1})]^j + [\MU_z \Vh_{t-1}]^j), 
\label{eq:z_igru} 
\\
 \tilde{\Vh}^j_t &= \phi([\MW \Ve(y_{t-1})]^j + [\MU(\Vr_t \odot 
\Vh_{t-1})]^j),  
\label{eq:th_irgu} \\
\Vh^j_t &= \Vz^j_t \Vh^j_{t-1} + (1-\Vz^j_t)\tilde{\Vh}_t^j \label{eq:h_igru},
\end{align}
where $\Vd_{i_t}$ is the output of the decoder. We can compute the probability 
of  each target character $y_t$ based on $\Vh_t$ with a softmax function:
\begin{equation}
  p(y_t | \{y_1, \ldots, y_{t-1} \}, \Vx) = \text{softmax}(\Vh_t). 
\label{eq:prediction}
\end{equation}

The current problem is that the number of outputs of decoder is much fewer than the 
target character sequence. It will be intractable by  conditionally picking outputs 
from the decoder when training in batch manner. In our model, we add a resampler 
to unfold the outputs of the decoder, which makes the training process more 
efficient.  Like the sampler, the resampler contains a $c\times t$ resampling 
matrix $\MR$, where $c$ is the number of delimiter of the target sequence and
$t$ is the length of the target. $\MR[i, j_1+1]$ to $\MR[i, j_2]$ are set as 1 if 
$j_1$ is the index of the $(i{-}1)$-th delimiter and $j_2$ is the index of the $i$-th 
delimiter in the target character sequence. The index of the $0$-th delimiter is set as 
$0$. For example, when the target output is 
``\underline{g} \underline{o} \underline{ } \underline{!} \underline{ }'' and 
the output of the decoder is $[\Vd_1, \Vd_2]$, the resample step will be:
\begin{align}
 [\Vd_1, \Vd_2]
 \begin{bmatrix}
    1      & 1 & 1 & 0 & 0 \\
    0      & 0  & 0 & 1 & 1
\end{bmatrix}
= [\Vd_1, \Vd_1, \Vd_1, \Vd_2, \Vd_2],  \nonumber 
\end{align}
therefore $\{\Vd_{i_1} , \Vd_{i_2} , \Vd_{i_3} , \Vd_{i_4} , \Vd_{i_5} \}$ is correspondingly
set to $\{\Vd_1, \Vd_1, \Vd_1, \Vd_2, \Vd_2\}$ in IGRU iterations. 
After resampling, we can compute  the probability of each target 
character by interpolator according to Eqns. \eqref{eq:seed_state} to 
\eqref{eq:prediction}. Compared with the two forward passes approach in 
\citep{luong2016achieving}, our resampling approach is more efficient.
Note that the resampling step is only necessary in 
training process. As explained in the Section \ref{subsec:gp}, the generation 
procedure is different from the training procedure.

\subsection{Model  Architectures}

\begin{figure}[!htb]
\centering
\includegraphics[width=0.9\textwidth]{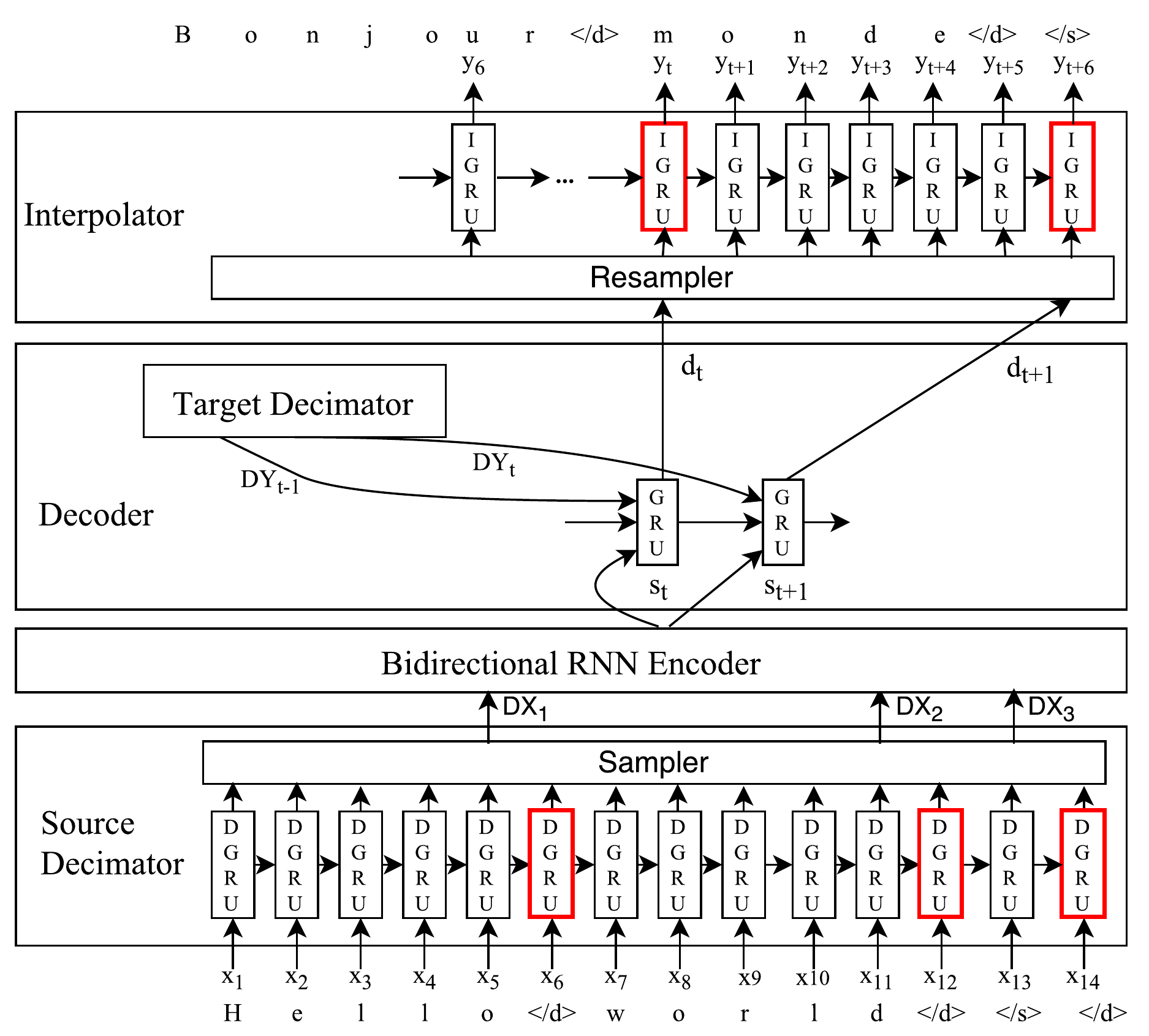}
\caption{Basic deep character-level neural machine translation. The DGRUs with 
red border indicate that the state is reset to the initial state and the IGRUs 
with red border indicate that the state should be set to the next output of the 
decoder. We refer readers to the supplementary material for a detailed 
illustration. }
\label{fig:dnmt}
\end{figure}

There are totally five recurrent neural networks in our model,  which can be 
divided into four layers as shown in  Figure \ref{fig:dnmt}. Figure 
\ref{fig:dnmt} 
illustrates the training procedure of a basic deep character-level neural 
machine translation. It is possible to use multi-layer 
recurrent neural networks to make the model deeper.
The first layer contains a source sequence decimator which samples the 
character level sequence according to  the delimiter (denoted as </d> in Figure 
\ref{fig:dnmt}). 
The second layer is a bidirectional RNN encoder which is identical to that of
\citep{bahdanau2014neural}. 
The third layer is the decoder. It takes the previous word representation as a 
feedback, which is produced by the target sequence decimator in our 
model. Then it combines the previous hidden state and the attention from the 
encoder to generate the next decoded vector.
The last layer  is the interpolator, which  takes the decoded vector 
as the state of IGRU. Based on the information of previous 
generated character,  the interpolator  generates the next character until generating an end of 
sentence token (denoted as </s> in Figure \ref{fig:dnmt}). 
With the help of  ``sampler'' and ``resampler,'' we can train our deep 
character-level neural translation model perfectly well in an end-to-end 
fashion.  

\subsection{Generation Procedure}
\label{subsec:gp}
We first encode the source sequence as  in the training procedure, then we generate 
the target sequence character by character based on the decoded vector $d_t$. 
Once we generate a delimiter, we should compute next decoded vector $d_{t+1}$ 
through the decoder by combining feedback $\text{DY}_{t}$ of the current 
generated word from the target decimator and the attention from the encoder. 
The generation procedure will terminate once an end of sentence 
(EOS) token is produced.

\section{Experiments}

We implement the model using Theano\citep{bergstra+al:2010-scipy, 
Bastien-Theano-2012} and Blocks \citep{van2015blocks}, the source code is available at github
\footnote{https://github.com/SwordYork/DCNMT}. We  train our model  on a GTX Titan X. 
For  fair comparison, we  evaluate our deep character-level neural machine translation
model  (DCNMT) on the  task of English-to-French translation, 
and conduct comparison with the basic word-level  neural machine translation model (RNNsearch) 
\citep{bahdanau2014neural} and  \citet{ling2015character}'s model (CBNMT).
We use  the same dataset as RNNsearch
which is the bilingual, parallel  corpora provided by ACL 
WMT'14\footnote{http://www.statmt.org/wmt14/translation-task.html}.

\subsection{Dataset}
The English-French parallel corpus of WMT'14 contains totaling 850M words. 
Following the same procedure of \citep{cho2014learning, bahdanau2014neural, 
axelrod2011domain}, we reduce the size of the corpus to 348M words. We use 
\emph{newstest2013} as the development set and evaluate the models on the 
\emph{newstest2014} which consists of 3003 sentences not present in the 
training data.

In terms of preprocessing, we only apply the usual tokenization. We 
choose a list of 120 most frequent characters  for each language which coveres 
nearly 100\%  of the training data. Those characters not included in the list are 
mapped to a special token (<unk>).

\subsection{Training Details}

We train two types of models with the sentences of length up to 50 words.
We follow \citep{bahdanau2014neural} to use similar hyperparameters. The encoder of 
both the models consists of forward and backward RNN, each has 1024 hidden units; 
and the decoder also contains 1024 hidden units. We use 30,000 most frequent 
words for RNNsearch and the word embedding dimensionality is 620.  We choose
120 most frequent characters for DCNMT and the character embedding 
dimensionality is 64. The DGRUs in both source decimator and target decimator  
have 512 hidden units. 
We test two models DCNMT-1 and DCNMT-2, which respectively contains  
a single-layer recurrent network and 2-layer recurrent network in the source decimator, bidirectional 
RNN encoder and decoder.

We use a stochastic gradient descent (SGD) with mini-batch of 80 sentences to 
train each model.  We calculate an adaptive step rate using Adadelta 
\citep{zeiler2012adadelta}.  For fair comparison, we trained each model for 
approximately two weeks.


We use  a beam search  to find a 
translation that approximately maximizes the 
conditional log-probability which is a commonly used approach in neural machine 
translation \citep{sutskever2014sequence, bahdanau2014neural}. In our DCNMT 
model, it is reasonable to search directly on character level to generate  a
translation.  

\section{Result and Analysis}
We show the comparison of quantitative results on the English-French 
translation task in Section \ref{sec:qresults}.  Apart from measuring 
translation quality, we analyze the 
efficiency of our model and effects of character-level modeling in more details.

\subsection{Quantitative Results} \label{sec:qresults}
We illustrate the efficiency of the deep character-level neural 
machine translation by comparing with the basic neural machine 
translation model (RNNsearch) and CBNMT \citep{ling2015character}.
We test on the  \emph{newstest2014} and measure the performance by BLEU score 
\citep{papineni2002bleu} which is commonly used in translation tasks. It is 
possible to further improve  the performance by using multi-layer deep 
RNNs, like the 4-layer LSTM used in \citep{sutskever2014sequence, 
luong2014addressing}.  
\begin{table}[h]
\centering
\caption{Model comparison}
\label{tb:model_c}
\begin{tabular}{lcccccc}
\hline
Model  & Parameters & Max len & Max mem & Updates per day & Epochs & BLEU
\\ \hline
RNNsearch  & 85.6M  &  50  & 9690MiB    & $\sim$ 58K      & 5.2  & 28.47
\\ \hline
CBNMT       & 34.8M   &  300  & 9680MiB   &  $\sim$ 54K/30K & 5.1  & 28.67
\\ \hline
DCNMT-1    & 33.6M     & 300 & 9627MiB  &  $\sim$ 42K   &  4.3 & 30.49
\\ \hline    
DCNMT-2    & 71.4M     & 300 & 11403MiB  &  $\sim$ 34K   &  3.8 & 31.76
\\ \hline
\end{tabular}
\end{table}

As shown in Table \ref{tb:model_c},  the number of parameters of RNNsearch is 
much large than character-based models and more than half of the parameters are in lookup 
table (55.8M) which is really redundant.  
CBNMT is layer-wise trained, thus the updates per day (54K) is similar to RNNsearch
when training attention, but the updates per day is dramatically reduced to 
30K when training the C2W component. Our models consume nearly the same memory as 
RNNsearch despite the 6 times longer sequence length.
It is commonly known that RNN is
less efficient than a simple lookup table, 
and there are 3 additional RNNs in DCNMT which decrease the update 
rate of DCNMT, however, still more efficient than CBNMT.
Compared with RNNsearch, both CBNMT and DCNMT achieves a higher BLEU score.
However, DCNMT-1 is more efficient than CBNMT and achieves much higher BLEU score 
using the similar amount of parameters, despite the use of IBM model 4 in CBNMT 
to produce the word alignments as a supervision.
Besides, the deeper model (DCNMT-2) 
further improves the performance within less epochs.


\subsection{Efficiency Analysis}
\begin{figure}[!htb]
\centering
 \subfigure[Training curve]
  { \label{sfig:lc} 
\includegraphics[width=0.49\linewidth]{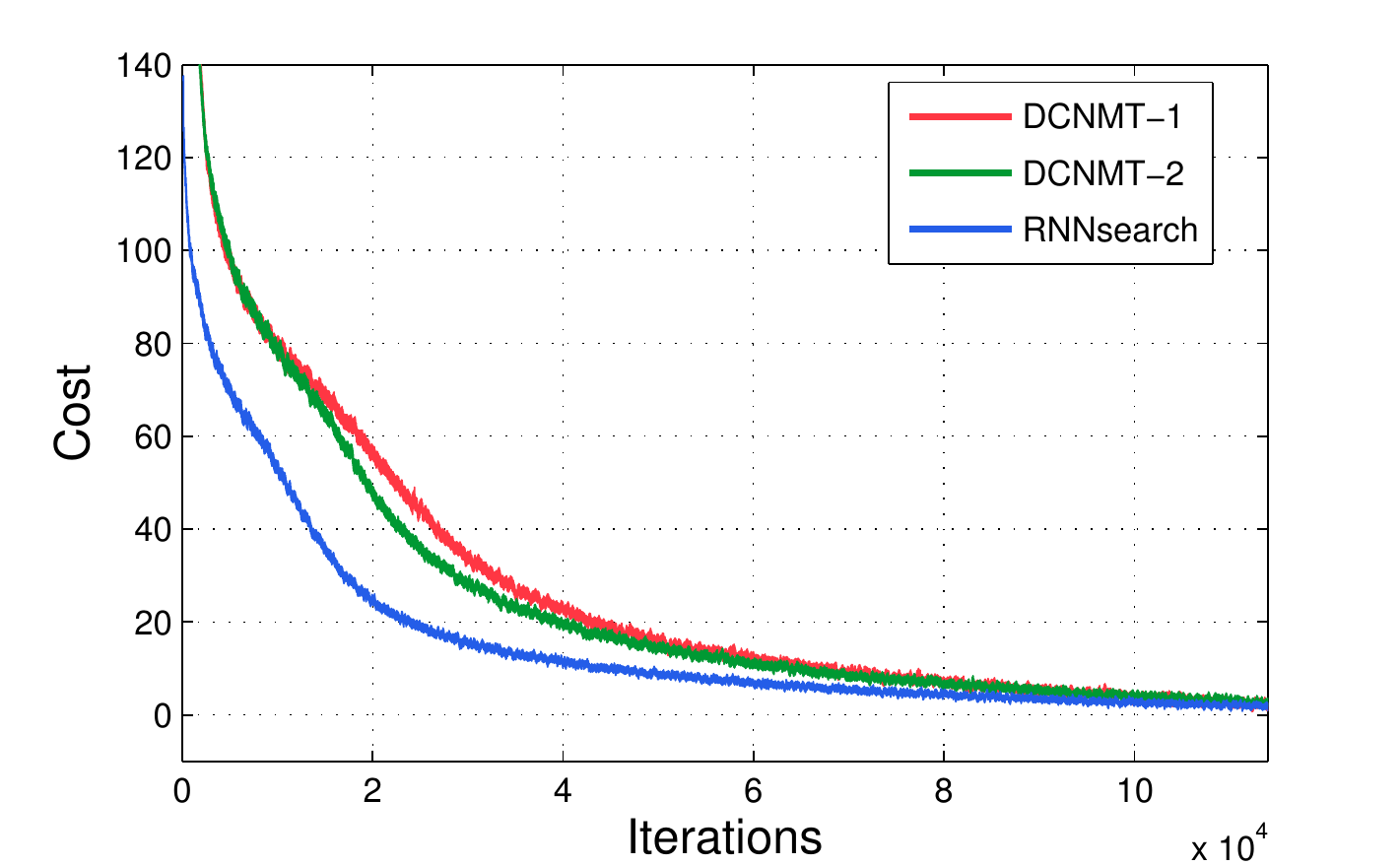}}
\subfigure[BLEU scores of first epoch]
{\label{sfig:bleuf}
\includegraphics[width=0.49\linewidth]{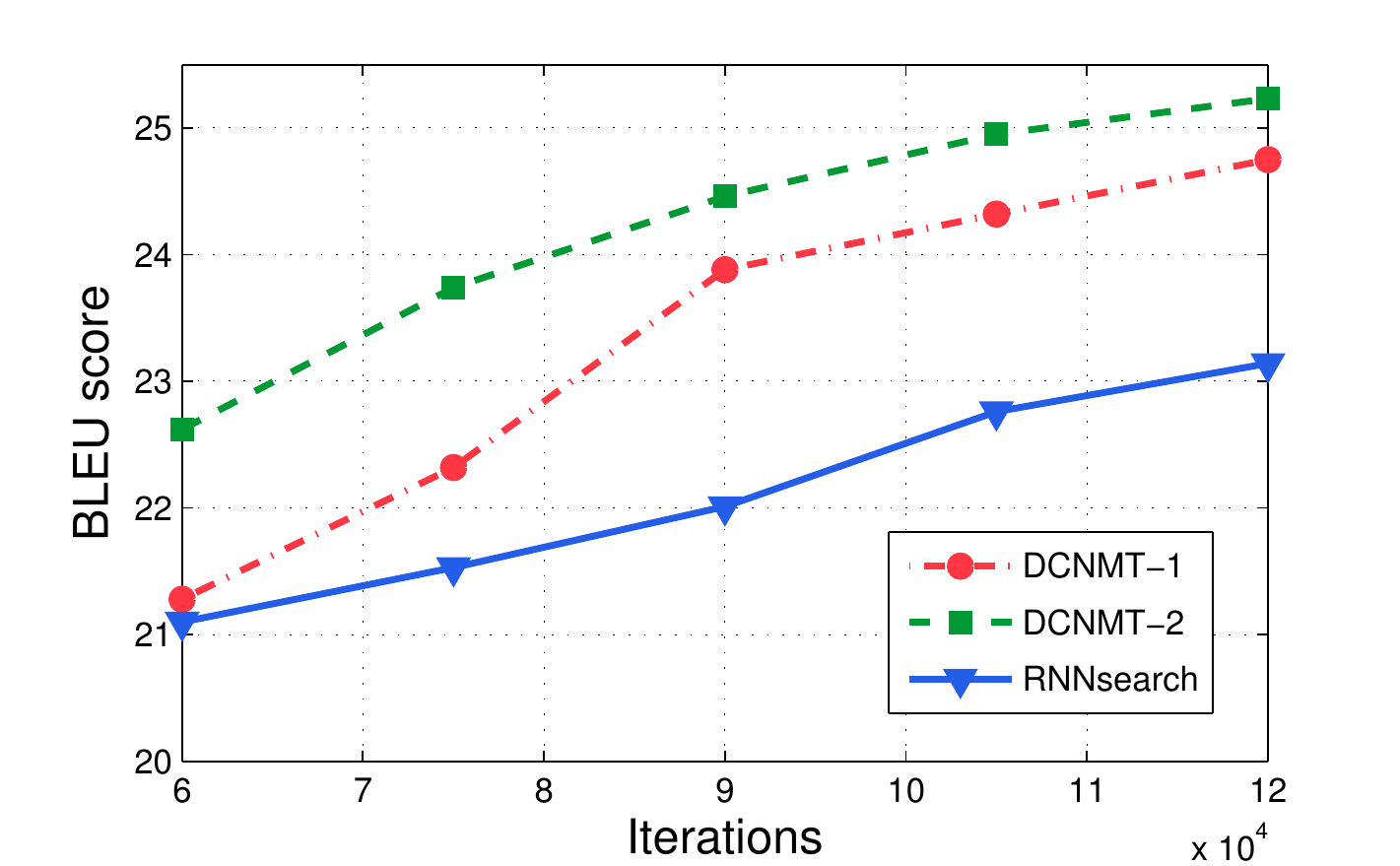}}
\caption{Traning process of RNNsearch and  DCNMT  }
\label{fig:ea}
\end{figure}

As shown in Figure \ref{sfig:lc}, the training curves of the three models are 
similar (we subtract the minimum cost value from cost for comparison). The 
DCNMT models start from a higher-cost state because they need to
learn one more abstraction compared with the word based RNNsearch.
Once they learnt some representation of words, 
they learn to translate just like the word-base models. Another evidence that 
 DCNMT and RNNsearch have similar behavior is the similar change of BLEU scores 
on development set as shown in Figure \ref{sfig:bleuf}. The DCNMT is able to 
perform as good as the RNNsearch when only trained by half of the epoch; in 
contrast to \citep{luong2016achieving, ling2015character},   they found that the 
character-level neural machine translation is extremely slow and difficult 
to train.  The comparison between DCNMT-1 and DCNMT-2 shows that the 
depth of recurrent network is critical for our model to achieve higher performance. 

It indicates that  DCNMT could outperform  the 
word-level  neural machine translation based on these analyses, and the 
out-of-vocabulary issue is solved at the cost of a small decrement of update 
rate.

\subsection{Translate Misspelled Words  and Word Embeddings}

Another advantage of our deep character-level neural machine translation is the 
ability to translate the misspelled words.  To the best of our knowledge, other 
extant neural machine translation models can not achieve this functionality. 
In  Table \ref{tb:trans_mis}, 
we list some examples where the source sentences are taken from 
\emph{newstest2013} but we change some words to misspelled words. We also list 
the translations from Google translate \footnote{The translations by Google 
translate were made on 20 May 2016.} and online demo of LISA

 \begin{table}[h]
\centering
\caption{Translation of misspelled words}
\label{tb:trans_mis}
\begin{tabular}{l|p{10cm}}
\hline
\multicolumn{2}{c}{Example 1}   \\ \hline
Source & Unlike in Canada, the American States are {\color{red}  
\emph{responisble}} for the  {\color{red}\emph{orgainisation}} of federal 
elections in the United States.
\\ \hline
Reference & Contrairement au Canada, les États américains sont    {\color{red}  
\emph{responsables}} de {\color{red}  \emph{l'organisation}} des élections 
fédérales aux États-Unis.
\\ \hline
Google translate& Contrairement au Canada , les États 
américains sont 
{\color{red}  \emph{responisble}} pour la {\color{red}  \emph{orgainisation}} 
des élections fédérales aux 
États-Unis.
\\ \hline
LISA &
Contrairement au Canada, les États-Unis sont  {\color{red}  \emph{UNK}} pour la 
 {\color{red}  \emph{UNK}} des élections fédérales aux États-Unis.
\\ \hline
DCNMT&   Contrairement au Canada , les États américains sont  {\color{red}  
\emph{responsables}} de {\color{red}  \emph{l'organisation}} des élections 
fédérales aux États-Unis.
\\ \hline 
\multicolumn{2}{c}{Example 2}   \\ \hline
Source & As a result, 180 bills restricting the {\color{red}  \emph{exrecise}} 
of the 
right to vote in 41 States were  {\color{red}  \emph{introuduced}} in 2011 
alone.
\\ \hline 
Reference & En conséquence , 180 projets de lois restreignant {\color{red}  
\emph{l'exercice}} du droit de vote dans 41 États furent  {\color{red}  
\emph{introduits}} durant la seule année de 2011.
\\ \hline
Google translate & En conséquence, 180 projets de loi restreignant la 
 {\color{red}  \emph{exrecise}} du droit de vote dans 41 Etats ont été 
 {\color{red}  \emph{introuduced}} en 2011 seulement . 
 \\ \hline
LISA & Par conséquent, 180 projets de loi restreignant le droit de vote dans 
41 États ont été  {\color{red}  \emph{UNK}} en 2011.
\\ \hline
DCNMT &   En conséquence, 180 projets de loi restreignant {\color{red}  
\emph{l'exercice}} du droit de vote dans 41 États ont été  
{\color{red} \emph{introduits}} en 2011 seuls. 
\\ \hline 
\end{tabular}
\end{table}

As listed in Table \ref{tb:trans_mis}, DCNMT is able to translate out the 
misspelled words correctly. For a word-based translator,  this is never possible 
because the misspelled words are mapped into <unk> token before training. Thus, 
it will produce an <unk> token or just take the word from source sentence as 
Google \citep{patent:20160117316} and many other models 
\citep{gulcehre2016pointing, luong2014addressing} 
do.

\begin{figure}[!htb]
\centering
 \subfigure[exercise and exrecise]
  { \label{sfig:exercice} 
\includegraphics[width=0.49\linewidth]{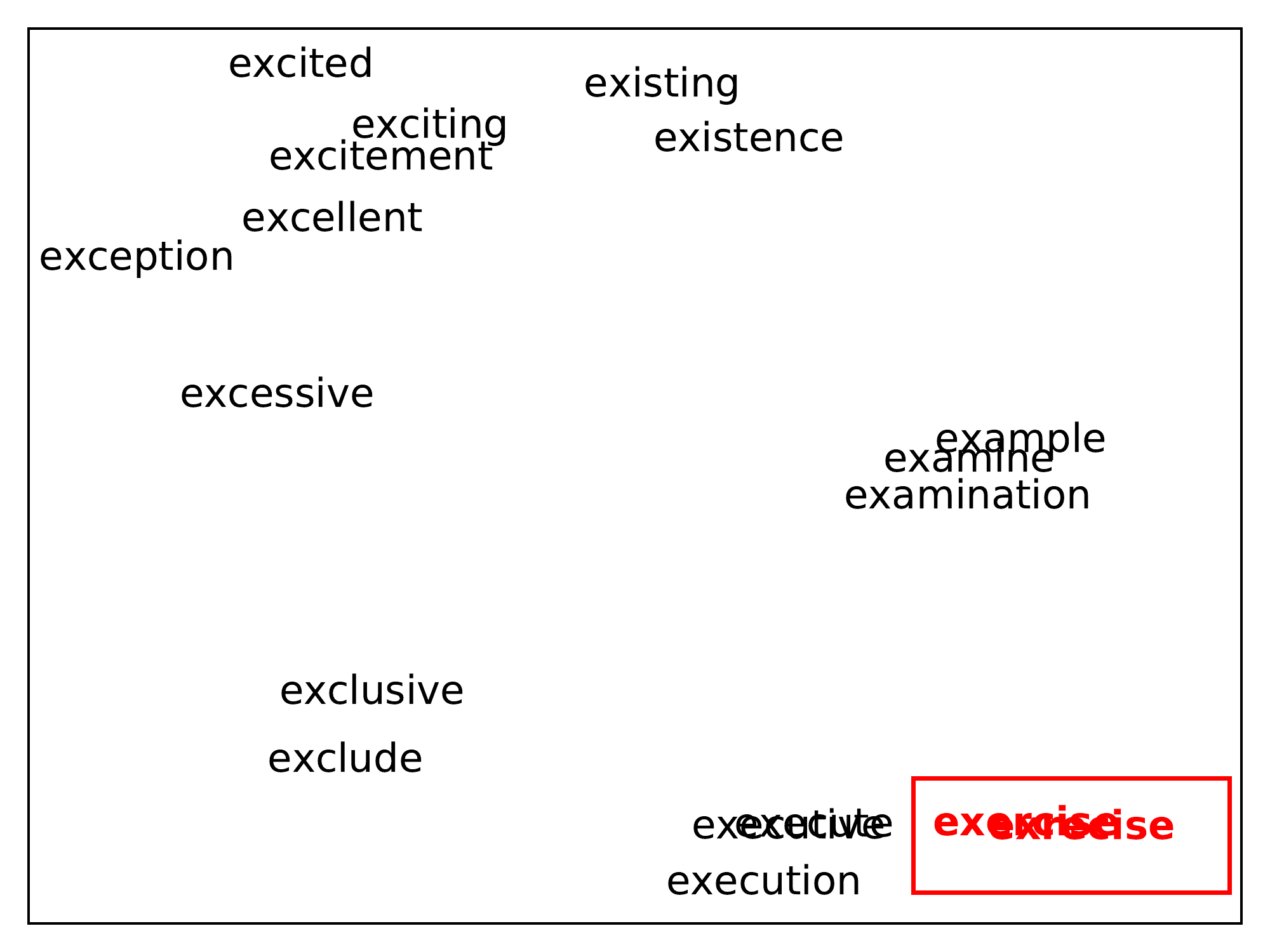}}
\subfigure[cluster of similar meaning words]
{\label{sfig:emb}
\includegraphics[width=0.49\linewidth]{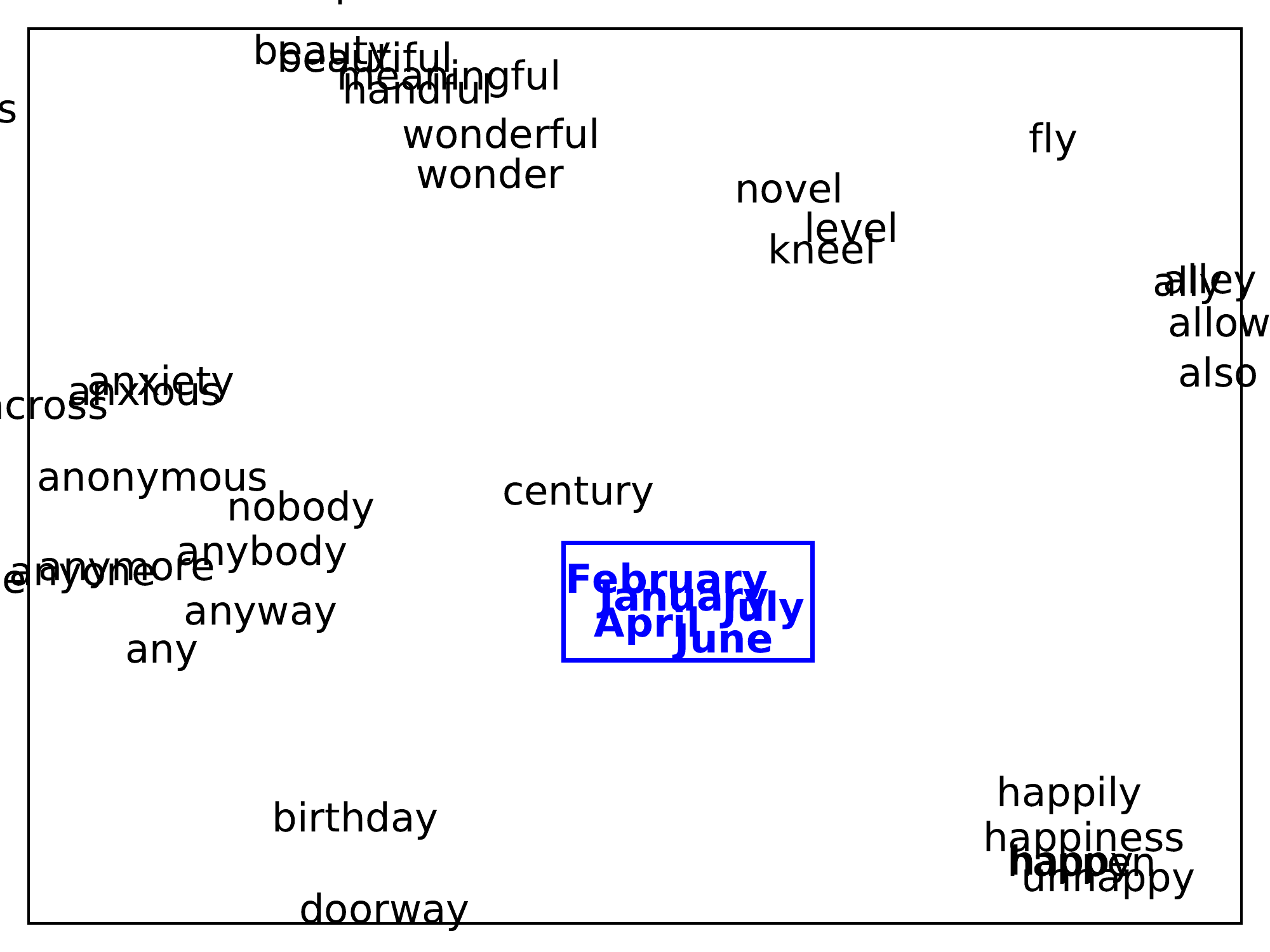}}
\caption{ t-SNE visualization of source word representations. We zoom in the 
particular parts for illustration.}
\label{fig:tsne}
\end{figure}

To investigate how  DCNMT works to translate the misspelled word, we
 visualize representation, produced by  DCNMT, of some frequent words in Figure 
\ref{fig:tsne}, which is computed by the t-SNE algorithm 
\citep{van2008visualizing}. We can conclude from Figure \ref{sfig:exercice} that 
words share a similar structure have a similar representation in DCNMT. The 
\emph{exercice} and \emph{exrecise} are extremely close as shown in  Figure 
\ref{sfig:exercice}.

Finally, the most surprising thing is that the words with similar meanings but different 
structure like  ``April,'' ``February,''  ``January,'' ``June'' and ``July'' 
 are clustered together as shown in Figure \ref{sfig:emb}.  It suggests that DCNMT is 
able to learn the embedding of the word in vector space to cluster words of 
similar meanings (usages) together like the word-level neural machine translation 
\citep{cho2014learning}.

\section{Conclusion}

In this paper we have proposed an efficient architecture to train the deep character-level neural machine 
translation model by introducing a \emph{decimator} and an \emph{interpolator}. 
We have demonstrated the efficiency of the training process and the effectiveness of 
the model  in comparison  with  the word-level models and \citet{ling2015character}'s model.
The higher BLEU score implies that  our deep character-level neural machine 
translation model  likely  outperforms the word-level models and the conventional character-based models.
It is  possible to further improve performance by using deeper GRU recurrent  networks or  training longer \citep{sutskever2014sequence}.

As a result of the character-level modeling, we have solved the 
out-of-vocabulary (OOV) issue that word-level models suffer from,  and  we have  obtained a 
new functionality that never achieved by extant neural machine translation model 
to translate the misspelled words.   More importantly,  the deep 
character-level is able to learn the similar embedding of the words with similar 
meanings  like the word-level models. Finally, it would be potentially possible that the idea behind  our approach could be  applied to many other tasks such as speech recognition and text classification.

\bibliography{reference}
\bibliographystyle{unsrtnat}

%
\end{document}